\documentclass{article}

\usepackage[arxiv]{optional} 
\opt{arxiv}{\usepackage{arxiv_iclr2019_conference}
\iclrfinalcopy 
}
\opt{icml}{\usepackage{icml2019}}

\usepackage{microtype}
\usepackage{graphicx}
\usepackage{subfigure}
\usepackage{booktabs} 
\usepackage{hyperref}
\usepackage{xspace}

\usepackage{amsmath,amsfonts,bm}

\def\eqref#1{equation~\ref{#1}}

\def\1{\bm{1}}

\def\vtheta{{\bm{\theta}}}

\DeclareMathAlphabet{\mathsfit}{\encodingdefault}{\sfdefault}{m}{sl}
\SetMathAlphabet{\mathsfit}{bold}{\encodingdefault}{\sfdefault}{bx}{n}

\usepackage{float}
\usepackage{url}
\usepackage{standalone}
\usepackage[capitalize]{cleveref}
\usepackage{multicol}
\usepackage{tikz}
\usepackage{amssymb}
\usepackage{algorithm}
\usepackage{algpseudocode}
\usepackage{wrapfig}
\usepackage{enumitem}
\usepackage{multirow}
\usepackage{booktabs,threeparttable}
  \newlist{inlinelist}{enumerate*}{1}
  \setlist*[inlinelist,1]{%
          label=(\roman*),
      }
      
\usetikzlibrary{calc}
\usetikzlibrary{positioning}
\usetikzlibrary{angles,quotes}
\usetikzlibrary{backgrounds}
\usetikzlibrary{external}
\usetikzlibrary{fit}
\usetikzlibrary{bayesnet}
\usetikzlibrary{calc}
\usetikzlibrary{fit}
\usetikzlibrary{spy, shadows}
\usetikzlibrary{arrows.meta}
\usetikzlibrary{shadings}
\usetikzlibrary{fadings}

\newcommand{\inputs}{\mathbf{X}}

\newcommand{\ie}{\textit{i.\,e., }}

\newcommand{\be}{\begin{equation}}
\newcommand{\ee}{\end{equation}}
\newcommand{\bea}{\begin{eqnarray}}
\newcommand{\eea}{\end{eqnarray}}
\newcommand{\beas}{\begin{eqnarray*}}
\newcommand{\eeas}{\end{eqnarray*}}

\newcommand{\given}{\,|\,}

\newcommand{\B}[1]{\bm{#1}}

\newcommand{\name}{PARSEC\xspace}

\opt{icml}{
\icmltitlerunning{Probabilistic Neural Architecture Search}

\begin{document}

\twocolumn[

\icmltitle{Probabilistic Neural Architecture Search}
\icmlsetsymbol{equal}{*}
\begin{icmlauthorlist}
\icmlauthor{Aeiau Zzzz}{equal,to}
\icmlauthor{Bauiu C.~Yyyy}{equal,to,goo}
\icmlauthor{Cieua Vvvvv}{goo}
\end{icmlauthorlist}
\icmlaffiliation{to}{Department of Computation, University of Torontoland, Torontoland, Canada}
\icmlaffiliation{goo}{Googol ShallowMind, New London, Michigan, USA}
\icmlcorrespondingauthor{Cieua Vvvvv}{c.vvvvv@googol.com}
\icmlcorrespondingauthor{Eee Pppp}{ep@eden.co.uk}
\icmlcorrespondingauthor{Eee Pppp}{ep@eden.co.uk}
\vskip 0.3in
]

\printAffiliationsAndNotice{\icmlEqualContribution}
}

\opt{arxiv}{
\title{Probabilistic Neural Architecture Search}
\author{Francesco Paolo Casale\thanks{These authors contributed equally to this work.}\\
Microsoft Research\\
Cambridge, MA, USA \\
\texttt{frcasale@microsoft.com} \\
\And
Jonathan Gordon\footnotemark[1] \\
University of Cambridge\\
Cambridge, UK \\
\texttt{jg801@cam.ac.uk} \\
\And
Nicol\'o Fusi\\
Microsoft Research\\
Cambridge, MA, USA \\
\texttt{fusi@microsoft.com} \\
}
\begin{document}
\maketitle
}

\begin{abstract}
In neural architecture search (NAS), the space of neural network architectures is automatically explored to maximize predictive accuracy for a given task.
Despite the success of recent approaches, most existing methods cannot be directly applied to large scale problems because of their prohibitive computational complexity or high memory usage.
In this work, we propose a Probabilistic approach to neural ARchitecture SEarCh (\name) that drastically reduces memory requirements while maintaining state-of-the-art computational complexity, making it possible to directly search over more complex architectures and larger datasets.
Our approach only requires as much memory as is needed to train a single architecture from our search space. This is due to a memory-efficient sampling procedure wherein we learn a probability distribution over high-performing neural network architectures.
Importantly, this framework enables us to transfer the distribution of architectures learnt on smaller problems to larger ones, further reducing the computational cost.
We showcase the advantages of our approach in applications to CIFAR-10 and ImageNet, where our approach outperforms methods with double its computational cost and matches the performance of methods with costs that are three orders of magnitude larger.
\end{abstract}

\section{Introduction}

Identifying a good neural network architecture for a given problem can be a difficult and time-consuming process, usually consisting in trying different combinations of layers and connection patterns until a "good" validation accuracy is finally reached. Due to the size of the optimization space, manually performing this complex optimization task can be daunting: one could choose different sizes or types of convolution, skip connections, layer sizes, number of layers, number of filters and so on. 

In recent years, there has been a surge of interest in automatically identifying neural network architectures, effectively replacing the expense of human time with the expense of computational time.
These approaches have often reached or exceeded the level of accuracy obtained by architectures that were tuned manually~\citep{zoph2017learning, cai2018path, liu2017progressive, zhong2018practical, zoph2016neural}.
Most of the early work in this area was focused on defining the search space (\emph{e.g.} which set of operation and connection patterns to consider) and the search method (\emph{e.g.} reinforcement learning, evolutionary approaches), and required to fully train and evaluate each architecture considered in the search.
Given that a typical search using these methods involved training hundreds to thousands of candidate architectures, the computational cost of these searches was in the order of hundreds to thousands of GPU-days.

A more recent line of work has focused on sharing weights across multiple architectures~\citep{pham2018efficient,cai2018efficient}.
This is typically achieved by defining an over-parameterized parent network containing all candidate paths in the architecture search space~\citep{liu2018darts, xie2018snas}.
In this setting, NAS reduces to estimate the importance of the candidate paths in the parent network and high-performing architectures can then be obtained by pruning unimportant paths in it.
While on one hand this approach drastically reduced the computational cost of NAS to a few GPU days, these methods have much higher memory requirements with respect to training an architecture of the search space, because of the much higher number of intermediate feature maps in the parent network.

The trade-off between high computational cost and high memory cost has resulted in the usage of different types of surrogates (or proxies) during architecture search. The main two types of surrogates are \emph{architecture surrogates} and \emph{dataset surrogates}. Architecture surrogates are small versions of the final network that are cheaper to store in memory and faster to train. For example, in image classification, this can be achieved by searching over the architecture of a recurrent convolutional unit, the \textit{cell}~\citep{
liu2017progressive, liu2017hierarchical, real2018regularized, cai2018path, liu2017learning, tan2018mnasnet, luo2018neural},
rather than directly on the global architecture.
Global networks with different complexities can then be obtained by stacking cells while varying the number of stacks, the number of filters, \textit{etc}. One advantage of this framework is that it allows to reuse cell architectures that are found for small global networks for larger ones.
The second type of surrogates, dataset surrogates, are datasets that can be used as a proxy to perform a quicker search.
For example, one could perform architecture search on CIFAR-10 and "transfer" the resulting architecture (with some modifications, e.g. changing the number of filters) to ImageNet~\citep{
liu2017progressive, liu2018darts, xie2018snas}.
While good performance on a surrogate task does not guarantee good performance on the final task (\emph{i.e.} fully-sized network on the target dataset), the vast majority of architecture search methods use at least one, if not both, of these surrogates.

In this work, we address these problems by casting neural network architecture search in a probabilistic modelling framework and propose a sampling-based optimization method to learn a probability distribution over
high-performing architectures for a specified supervised task. Our probabilistic framework, called \name (Probabilistic neural ARchitecture SEarCh), has multiple advantages over existing NAS methods:

\begin{itemize}
\item our search procedure is memory-efficient, as only the feature maps associated with the sampled architectures
are loaded on the GPU. This allows us to directly search over fully-sized architectures and larger datasets;
\item
because of \name's search space and probabilistic foundations,
our approach can transfer probability distributions over architectures learnt on small surrogates to larger networks and datasets, enabling us to further reduce the computational cost;
\item our search procedure is computationally efficient and can be run in less than a day on a single GPU on CIFAR-10;
\item
in experiments on CIFAR-10 and ImageNet, we show that \name outperforms other methods that consider the same architecture search space, while drastically reducing the search time.
When evaluating architectures on ImageNet, the best architecture found by \name on CIFAR-10 in less than one GPU-day outperforms architectures identified by other methods with
comparable computational cost, while matching the accuracy of architectures found by methods with computational costs that are two to three orders of magnitude larger.

\end{itemize}

\section{Related Work}
\label{sec:background}

Neural architecture search (NAS) is a line of research that has received significant attention over the last few years.
The goal of NAS methods is to search for architectures that have good performances on a specific task.
This is achieved by performing a heuristic search in a predefined architecture search space.

For image classification tasks, most works define a search space in terms of \textit{cells}, computational graphs of neural primitives (\emph{e.g.}, convolution or pooling operations) that can then be stacked to compose global networks \citep{zoph2016neural, liu2017progressive}.
See Figure 1a for an illustration.

Multiple search methods have been considered to traverse the space of architectures, including reinforcement learning (RL; \citet{zoph2016neural, baker2016designing}), genetic algorithms \citep{liu2017hierarchical, xie2017genetic, zoph2017learning}, and progressive search methods \citep{liu2017progressive, negrinho2017deeparchitect}.
A significant drawback of these methods is that they require fully training and evaluating hundreds (or thousands) of intermediate models during search, resulting in searches that require thousands of GPU compute hours.

More recently, researchers have observed that the key source of computational overhead is the need to fully train intermediate models during search~\citep{pham2018efficient, cai2018efficient}.
This observation lead to the idea of \textit{weight sharing} during training~\citep{pham2018efficient}. Here intermediate models all share the same weights, which are updated during search iterations. In these works, the entire search space is viewed as a single, large parent network containing all possible paths, while instances of architectures are child networks from the parent. 
Within this framework, \citet{pham2018efficient} introduce a controller on the large graph that activates subsets of the graph at each iteration.
The controller is trained with policy gradient to discover subgraphs with good validation accuracy. 
\citet{liu2018darts} propose instead a continuous relaxation to the exact problem, where hard selections on the paths to select in the parent network are replaced by continuous parameters that weight the importance of each path.
The weights are tuned with gradient descent on held out validation cross-entropy, and at the end of search a heuristic is employed to extract a single graph based on the finalized weights.
\citet{xie2018snas} represent the search space of with a set of one-hot random variables in a fully factorized joint distribution.
All these papers produce compelling results in a 1000x reduction in compute time with respect to previous method, relying either on architecture surrogates or on dataset surrogates (or both).

Our work is most closely related to DARTS \citep{liu2018darts} and SNAS \citep{xie2018snas}.
We use the same search space and computational graph as DARTS, but rather than employing a continuous relaxation, we directly tackle the optimization problem by viewing search evaluations as samples from an underlying distribution over architectures, a procedure that is similar to SNAS~\citep{xie2018snas}.
In contrast to SNAS, we consider a different factorization of the probability distribution over architectures that leads to a dramatic reduction of the memory requirements: the memory requirements of \name are the same as needed for training a single architecture of the search space.
Also, we propose leveraging previous searches as prior information to speed up convergence on a new dataset or different global architecures.

Finally, another related method is ProxylessNAS \citep{cai2018proxylessnas},
which achieves low memory requirements by applying a binary mask to the architecture topology and operations, forcing only a path through the network to be active and thus loaded in GPU memory.
The parameters of the binary mask are learned using BinaryConnect \citep{courbariaux2015binaryconnect}.
ProxylessNAS is the first method not requiring surrogates for architecture search.
However, in contrast to \name, ProxylessNAS
is used for searching over global architectures rather than over a recurrent unit (\ie the cell) as done here, thereby requiring a strong prior knowledge on which global network search space should be selected for specific datasets. This requirement also prevents transferring knowledge across related tasks.
In contrast, our goal in this work is to learn a set of good cells that perform well across different architecture sizes and datasets.
Leveraging similarity between problems, we can then efficiently fine-tune these architecture choices to perform well for specific architecture sizes and datasets.

\section{Methodology}
\label{sec:model}

In this section, we present our probabilistic approach to neural architecture search.
In Section \ref{sec:space} we describe the architecture search space considered in our work;
in Section \ref{sec:deterministic} we summarize the deterministic NAS method proposed in \citet{liu2018darts}, our model builds upon;
in Section \ref{sec:probabilistic} we introduce our probabilistic framework; finally, in Section \ref{sec:iw-mceb} we give full algorithmic details of our architecture search procedure.

\subsection{Architecture Search Space}
\label{sec:space}
\begin{figure*}[h]
  \centering
  \subfigure[Complete network]{\includegraphics[width=.28\linewidth]{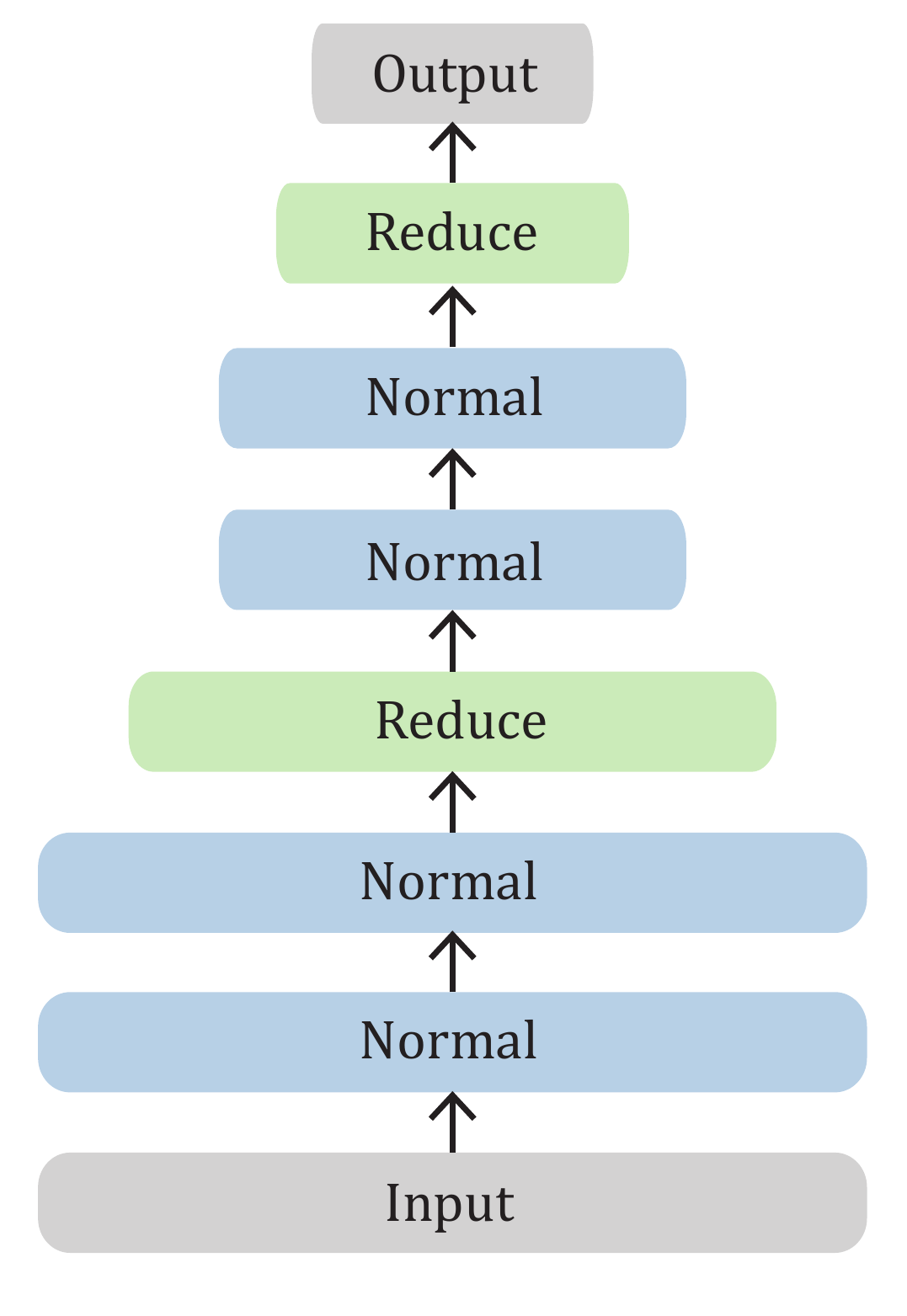}}\quad
  \subfigure[Normal cell]{\includegraphics[width=.33\linewidth]{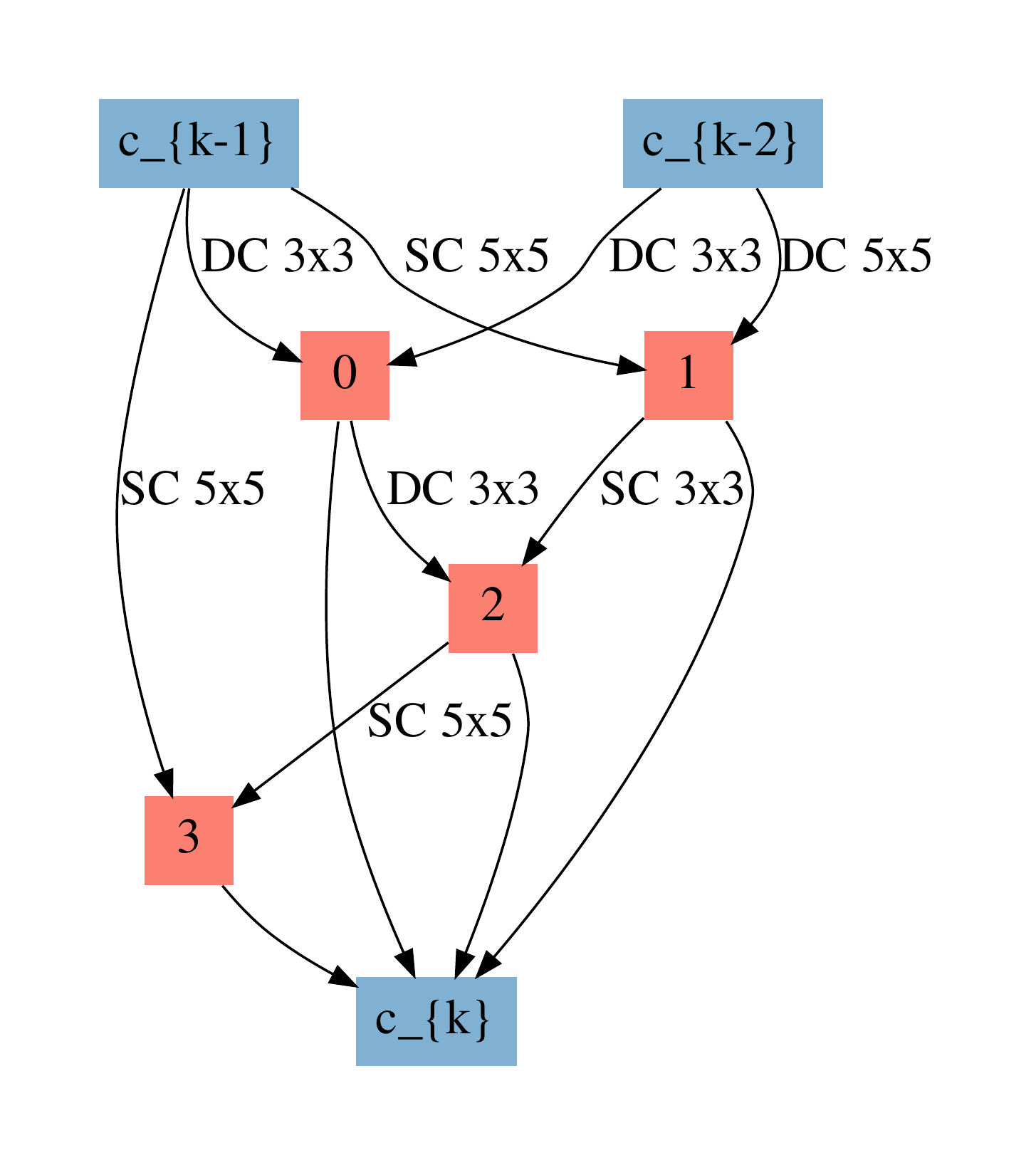}}\quad
  \subfigure[Reduction cell]{\includegraphics[width=.33\linewidth]{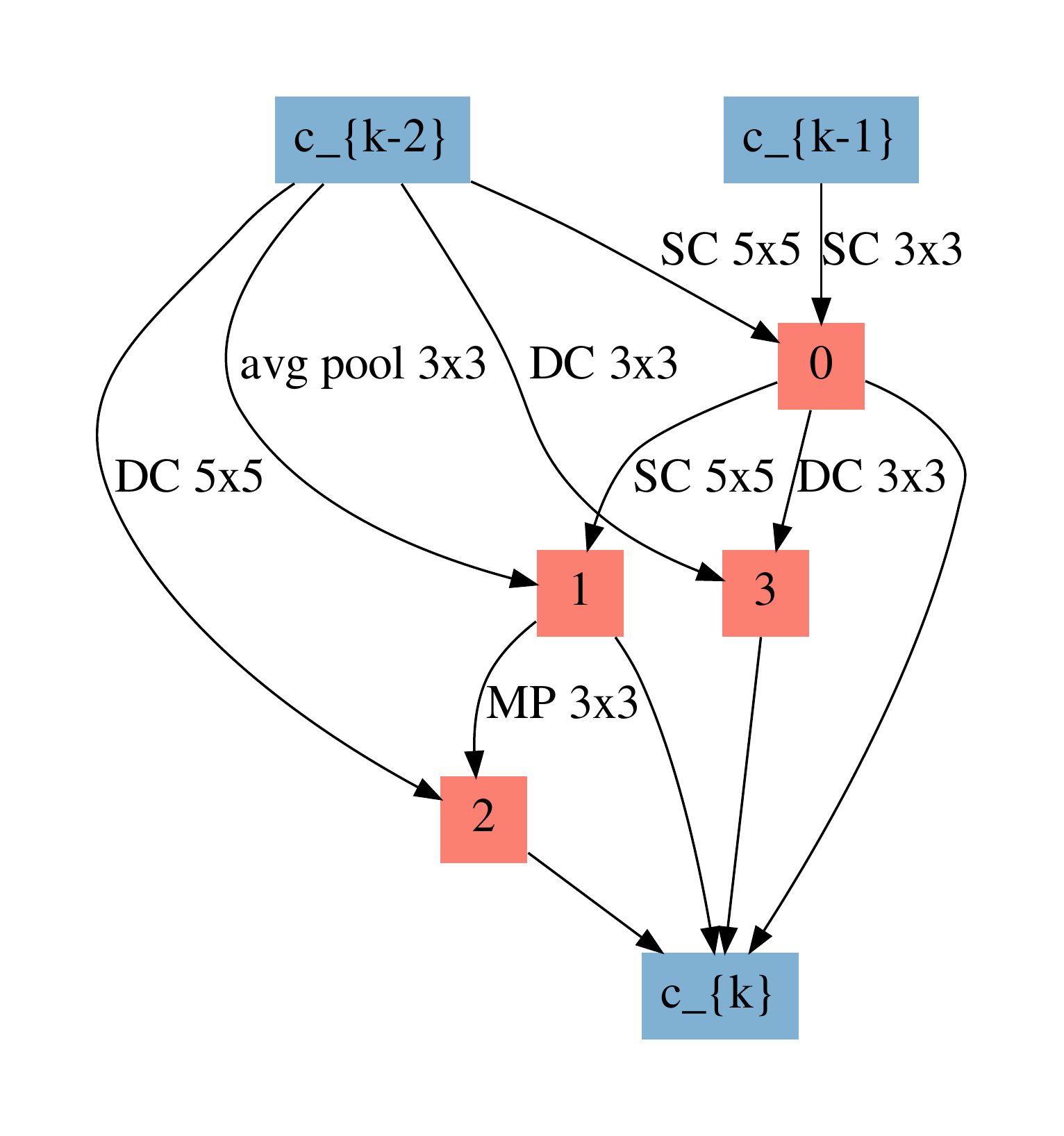}}
\caption{(a) Global network obtained stacking normal and reduction search.
(b) Example of normal and (c) reduction cells.}
\label{fig:cell2arch}
\end{figure*}
In this work, we consider the same architecture search space as in DARTS~\citep{liu2018darts}.
Neural network architectures are obtained by stacking a recurrent convolutional unit, denoted as \emph{cell}~\citep{zoph2017learning}.
A cell is defined as a directed acyclic graph of $N$ ordered nodes, $\{\B{z}_1, \dots, \B{z}_N\}$, taking the output of the two previous cells as inputs and returning the concatenation of all its intermediate nodes as output.
An intermediate node, $\B{z}_k$, is defined as the sum of two of the previous nodes, $\B{z}_i$ and $\B{z}_j$ with $i,j<k$, after applying primitive operations $o_{j,k}$ and $o_{j,k}$, respectively:
\begin{equation}
    \label{eqn:node_computation}
    \B{z}_k = o_{i,k}(\B{z}_i)+o_{j,k}(\B{z}_j),\;\;\;i<k,\;\;\;j<k.
\end{equation}
A node is thus fully specified by the two previous nodes selected as inputs and the two primitive operations that are applied to either of them.
As in~\citet{liu2018darts}, we consider the following $P=7$ primitive operations:
\begin{itemize}
\item identity;
\item $3\times3$ average pooling;
\item $3\times3$ max pooling;
\item $3\times3$ separable convolutions;
\item $5\times5$ separable convolutions;
\item $3\times3$ dilated separable convolutions;
\item $5\times5$ dilated separable convolutions.
\end{itemize}
All convolutions are preceded by a ReLu transformation~\citep{xu2015empirical} and followed by batch normalization \citep{ioffe2015batch}, and padding is added to preserve spatial dimensions of feature maps. 
A cell is fully specified when the inputs and operations of each intermediate node are defined.

For a cell with $N=4$ nodes, this corresponds to $\prod_{n=1}^NP^2\times(n+1)^2\approx{10^{10}}$ possible choices (see ~\citet{liu2017progressive}).
\ifx 0
\begin{eqnarray}
\underbrace{P^2\times2^2}_{ \nonumber
\text{node 1}}\times\underbrace{P^2\times3^2}_{
\text{node 2}}\times\underbrace{P^2\times4^2}_{
\text{node 3}}\times\underbrace{P^2\times5^2}_{
\text{node 4}}\approx10^{10}
\end{eqnarray} 
choices.
\fi

Finally, following previous work~\citep{zoph2016neural,liu2018darts}, we also consider two different types of cells: \emph{normal} cells and \emph{reduction} cells.
Operations in a \emph{normal} cell have stride one and do not change the shape of the feature maps, whereas operations in a \emph{reduction} cell have stride two and halve the spatial dimensions of the feature maps.
When stacking cells, every third cell is a reduction cell, and the rest are normal (Figure \ref{fig:cell2arch}a).
The architectures of these two cells may be different and are learned jointly during search (Figure \ref{fig:cell2arch}b-c shows an example of normal and reduction cell, respectively).

\subsection{Deterministic Approach to Architecture Search}
\label{sec:deterministic}

DARTS~\citep{liu2018darts} is a deterministic approach to neural architecture search that uses a continuous relaxation on the categorical choices of inputs and operations, so that the architecture can be optimized using gradient descent.
Specifically, \citet{liu2018darts} consider an over-parametrized parent network containing all possible paths between nodes:
\begin{eqnarray}
\B{z}_k = \sum_{i,p} \alpha_{ik}^{p}\cdot o^p_{i,k}(\B{z}_i),
\end{eqnarray}
where indices $i$ and $p$ run over all possible inputs and operations for node $k$ respectively, $o^p_{i,k}$ denotes primitive operation $p$ on input node $i$ to contribute to output node $k$, and $\alpha^p_{i,k}$ is the weight of the path.
The path weights $\alpha^p_{i,k}$ of all operations between two nodes are set to sum up to 1, \ie $\sum_p{\alpha^p_{i,k}}=1$.
In order to enable the automatic pruning of unimportant input nodes during search, a null operation is added to the set of primitive operations.
After this relaxation, architecture search reduces to an optimization problem over the continuous architecture parameters $\B{\alpha}=\{\alpha_{i,k}^p\}$, which measure the importance of the different paths in the parent network.
Specifically, the authors propose to jointly optimize the network weights $\B{v}$ of the parent (\ie~the union of the weights associated to each operation $o_{i,k}^p$) and the architecture parameters $\B{\alpha}$ using gradient descent on different splits of the data, namely a train and a search set.
After the search, an architecture of the original search space is obtained by (i) selecting for each node the two strongest input nodes and (ii) selecting the strongest operation at each remaining edge.
The obtained architecture is then trained from scratch (\ie~weights are re-initialized) for the supervised task at hand.
For further details on the optimization procedure, we refer to the original DARTS paper~\citep{liu2018darts}.

\subsection{Probabilistic Neural Architecture Search}
\label{sec:probabilistic}
\begin{figure*}[h]
	\begin{center}
		\includegraphics[width=1.01\textwidth]{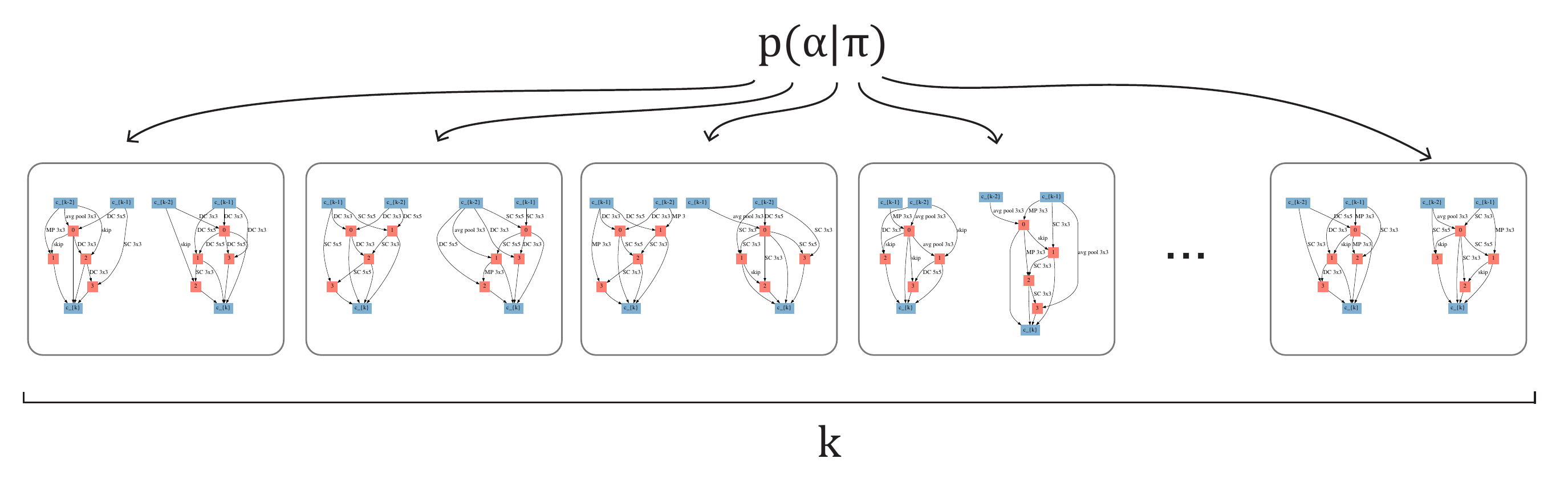}
	\end{center}
\caption{
Pictorial representation of architecture sampling. Each sample consists of a normal and a reduction cell, which are shown side-to-side.}
\label{fig:cell2arch}
\end{figure*}

As opposed to the deterministic framework in DARTS, we consider a Probabilistic Approach to Neural ARchitecture SEarCh (\name).
We introduce a prior $p(\B{\alpha}|\B{\pi})$ on the choices of inputs and operations that define the cell (Figure 3a), where hyper-parameters $\B{\pi}$ are the probabilities corresponding to the different choices.
In this framework, child networks from the search space are directly obtained as samples from the specified architecture distribution, and thus neural architecture is reduced to inferring the probability distribution $p(\B{\alpha}|\B{\pi})$ of high-performing architectures on the task at hand.
Given a supervised task and denoting with $\mathbf{y}$ the targets and with $\mathbf{X}$ the input features, this is achieved by optimizing the continuous prior hyper-parameters $\B{\pi}$ through an empirical Bayes Monte Carlo procedure~\citep{carlin2000empirical}.
Specifically, we optimize:
\begin{equation}
p(\B{y}\given \inputs,\B{v},\B{\pi}) =
\int p(\B{y}| \inputs,\B{v}, \B{\alpha}) p(\B{\alpha} | \B{\pi})\mathrm{d}\B{\alpha},
\end{equation}
with respect to the network weights $\B{v}$, which are shared across all child architectures (\ie samples), and the prior hyper-parameters $\B{\pi}$, \ie~the architecture parameters.

\paragraph{Architecture distribution.}

In the search space we introduced in Section \ref{sec:space}, each intermediate node is specified by defining (i) the two input predecessor nodes and (ii) the two primitive operations to apply to each of the input nodes.
We assume independent categorical distributions over each input/operation pair.
Specifically, for normal and reduction cells we set
\begin{eqnarray}
p(\B{\alpha}^\text{(n)}|\B{\pi}^\text{(n)})&=&\prod_{n=1}^N\prod_{i=1}^2\text{Cat}(\B{\alpha}_{n,i}^\text{(n)}|\B{\pi}_{n,i}^\text{(n)}),
\\
p(\B{\alpha}^\text{(r)}|\B{\pi}^\text{(r)})&=&\prod_{n=1}^N\prod_{i=1}^2\text{Cat}(\B{\alpha}_{n,i}^\text{(r)}|\B{\pi}_{n,i}^\text{(r)}),
\label{eq:probpnas}
\end{eqnarray}
where $n$ runs over the nodes, $i$ runs over the inputs of each node (in our search space each node has two inputs),
$\B{\pi}_{n,i}^\text{(n)}$ and $\B{\pi}_{n,i}^\text{(r)}$ are the vectors of probabilities for all possible incoming node/operation pairs and finally, we introduced
$\B{\alpha}^{(\cdot)}=\{\B{\alpha}_{n,i}^{(\cdot)}\}$ and
$\B{\pi}^{(\cdot)}=\{\B{\pi}_{n,i}^{(\cdot)}\}$.
Note that each sample from this distribution is a child architecture from the architecture search space.
As in our search procedure all computations are done on child networks sampled from the architecture distribution (as explained in detail in the next section), the resulting NAS algorithm has the same memory requirements of training a single architecture from the search space. This is not the case, for example, for either SNAS or DARTS.

\subsection{Importance-Weighted Monte Carlo empirical Bayes}
\label{sec:iw-mceb}

In this section, we develop an importance weighted EB procedure for jointly optimizing $\B{\pi}$ and $\B{v}$. We begin by introducing the following estimator:
\begin{equation}
    \label{eqn:py_iw_estimator}
    \begin{split}
    p(\B{y} | \inputs, \B{v}, \B{\pi}) & = \int p(\B{y} | \inputs, \B{v}, \B{\alpha}) p(\B{\alpha} | \B{\pi}) \mathrm{d}\B{\alpha} \\   
    & \approx \frac{1}{K}\sum_{k} p(\B{y} | \inputs, \B{v}, \B{\alpha}_k), 
   \end{split}
\end{equation}
with $\B{\alpha}_k \sim p(\B{\alpha} | \B{\pi})$.
Note that the gradients can be written as
{\footnotesize
\begin{eqnarray}
\nabla_{\B{v},\B{\pi}}\log p(\B{y}| \B{X}, \B{v}, \B{\pi})
&=&
\frac{1}{p(\B{y}| \B{X}, \B{v}, \B{\pi})}
\int \nabla_{\B{v},\B{\pi}} p(\B{y}, \B{\alpha}| \B{X}, \B{v}, \B{\pi}) \mathrm{d}\alpha
\label{eqn:gradient_manipulation}\\
&=&\frac{1}{p(\B{y}| \B{X}, \B{v}, \B{\pi})}
\int \nabla_{\B{v},\B{\pi}} \log p(\B{y}, \B{\alpha}| \B{X}, \B{v}, \B{\pi}) p(\B{y}, \B{\alpha}| \B{X}, \B{v}, \B{\pi})\mathrm{d}\B{\alpha}
\nonumber
\end{eqnarray}
}
Finally, plugging the estimator in \cref{eqn:py_iw_estimator} into \cref{eqn:gradient_manipulation} and taking an importance sampling estimator to the expectation, we obtain tractable gradient estimators for $\displaystyle \vtheta$ and $\bm \pi$:
{\footnotesize
\begin{eqnarray}
    \nabla_{\B{v}}\log p(\B{y} | \inputs, \B{v}, \B{\pi}) &\approx&
    \sum_{k=1}^K \tilde{\omega}_k \nabla_{\B{v}}\log p(\B{y} | \inputs, \B{v}, \B{\alpha}_k)\triangleq \tilde{\nabla}_{\B{v}}, \nonumber\\ 
    \nabla_{\B{\pi}}\log p(\B{y} | \inputs, \B{v}, \B{\pi}) &\approx&
    \sum_{k=1}^K \tilde{\omega}_k \nabla_{\B{\pi}}\log p(\B{\alpha}_k | \B{\pi})\triangleq \tilde{\nabla}_{\bm \pi}, \label{eqn:grad_estimators}
\end{eqnarray}
}
where
$\B{\alpha}_k \sim p(\B{\alpha} | \B{\pi})$
and
$\tilde{\omega}_k = \frac{p(\B{y} | \inputs, \B{v}, \B{\alpha}_k)}{\sum_j p(\B{y} | \inputs, \B{v}, \B{\alpha}_j)}$.

Note that the gradient estimators in \cref{eqn:grad_estimators} are weighted average of the gradients computed on $K$ child networks sampled from $p(\B{\alpha}|\B{\pi})$, where the importance weights $\tilde{\omega}_k$ are proportional to the model likelihood of child network $k$, $p(\B{y} | \inputs, \B{v}, \B{\alpha}_k)$.
The update rules and procedure derived here are similar to the ones used in the reweighted wake-sleep procedure~\citep{bornschein2014reweighted, le2018revisiting}, where rather than updating variational parameters of an approximate posterior distribution we are updating the parameters of the prior.
Following the approach taken in DARTS~\citep{liu2018darts}, in order to avoid over-fitting during the joint optimization of the architecture parameters and the shared weights, we consider different splits of the data (specifically, a \emph{search} and a \emph{train} sets) for the estimation of the importance weights and the estimation of the network weight updates. A full description of the derived importance weighted Monte-Carlo EB discussed in this section is given in \cref{alg:iwmceb}.

After joint optimization of the network weights $\B{v}$ and the architecture parameters $\B{\pi}$, one can extract a specific configuration of hyper-parameters in several ways.
In this work, we consider the mode of the architecture distribution and train it from scratch.
An alternative strategy is to train multiple samples from the learnt distribution from scratch and consider model ensembling.
We plan to explore this direction in future work.

\begin{algorithm}[t]
	\caption{Importance weighted Monte-Carlo EB algorithm used for joint training of the network and architecture parameters.}
	\begin{algorithmic}[1]
	    \State{Define train and search splits of the data}
	    \State{$\B{\theta}, \B{\pi} \leftarrow$ Initial values}
		\State{Initialize $\B{\theta}$ and $\B{\pi}$}
		\Repeat{}
		\State{Sample batch $(\inputs^\text{(s)}, \B{y}^\text{(s)})$ from search set}
		\For{$k \in \{1,\dots,K\}$}
		\State{Sample architecture $\B{\alpha}_k$ from $p(\B{\alpha} | \B{\pi})$}
		\State{$\omega_k\leftarrow p(\B{y}^\text{(s)} | \inputs^\text{(s)}, \B{v}, \B{\alpha})$}
		\EndFor
		\State{Compute normalized weights $\tilde{\omega}_k=\frac{\omega_k}{\sum_{j=1}^K\omega}_j$}
		\State{Sample batch $(\inputs^\text{(s)}, \B{y}^\text{(t)})$ from train set}
		\For{$k \in \{1,\dots,K\}$}
		\State{
		$g_{\B{v}, k}
		\leftarrow
		\nabla_{\B{v}}\log p(\B{y}^\text{(t)} | \inputs^\text{(t)}, \B{v}, \B{\alpha}_k)$
		}
		\State{
		$g_{\B{\pi}, k}
		\leftarrow
		\nabla_{\B{\pi}}\log p(\B{\alpha}_k | \B{\pi})$
		}
		\EndFor
		\State{Update $\B{v}$ by $\tilde{\nabla}_{\B{v}}=\sum_k\tilde{\omega}_kg_{\B{w}, k}$
		}
		\State{Update $\B{\pi}$ by $\tilde{\nabla}_{\B{\pi}}=\sum_k\tilde{\omega}_kg_{\B{\pi}, k}$
		}
		\Until{Number of epochs is reached or convergence is achieved}
		\State {\textbf{Return:} $\left(\displaystyle \vtheta, \pi\right)$}
	\end{algorithmic}
	\label{alg:iwmceb}
\end{algorithm}

\section{Experiments and Results}
\label{sec:results}

We validate our method by performing architecture search using CIFAR-10 and evaluating performance both on CIFAR-10 and ImageNet.
In Section \ref{sec:results_cifar_small}, we evaluate the performance of our approach using the same procedure considered in \citep{liu2018darts}:
(i) search a good cell architecture using a small global networks,
(ii) evaluate the best cell architecture on a fully-sized global network.
In Section \ref{sec:results_cifar_big}, we use our probabilistic framework to tune the distribution over architectures identified using the small global network using the fully-sized global network and show that this further improves performance.
Finally, in Section \ref{sec:results_imagenet} we evaluate the networks searched on CIFAR-10 using ImageNet. 

Overall, our results show that:
\begin{enumerate}
    \item when searching over small network surrogates, our method matches or exceeds (dependending on the dataset) the performance of methods with similar search space and comparable computational cost.
    \item fine-tuning without network surrogates on CIFAR-10 (\ie directly operating over fully-sized networks) further improves the results in both datasets.
\end{enumerate}

\subsection{Neural architecture search using CIFAR-10}
\label{sec:results_cifar_small}

We start by evaluating the performance of our approach in the same exact setting as DARTS \citep{liu2018darts}.
Specifically, we first search good cell architectures using a global small network, then we evaluate the best cell when using a fully-sized global network. 

To perform the search we hold out $50\%$ of the CIFAR-10 training set and use it as a search set, \ie we use it to compute the architecture importance weights (Algorithm 1).
We learn the architecture parameters using Adam~\citep{kingma2014adam} with learning rate 0.02 and $\mathbf{\beta} = (0.5, 0.999)$ for 100 epochs with $k=16$.
All other parameter and settings are identical to the ones used in \citep{liu2018darts}.
Note that in \name, $k$ acts as a fidelity parameter: higher $k$ will result in 
more exploration of the architecture space at the expense of higher computational costs.

\begin{figure*}[t]
  \centering
  \subfigure[Validation accuracy of the mode of the architecture distribution vs search epochs]{\includegraphics[width=.43\linewidth]{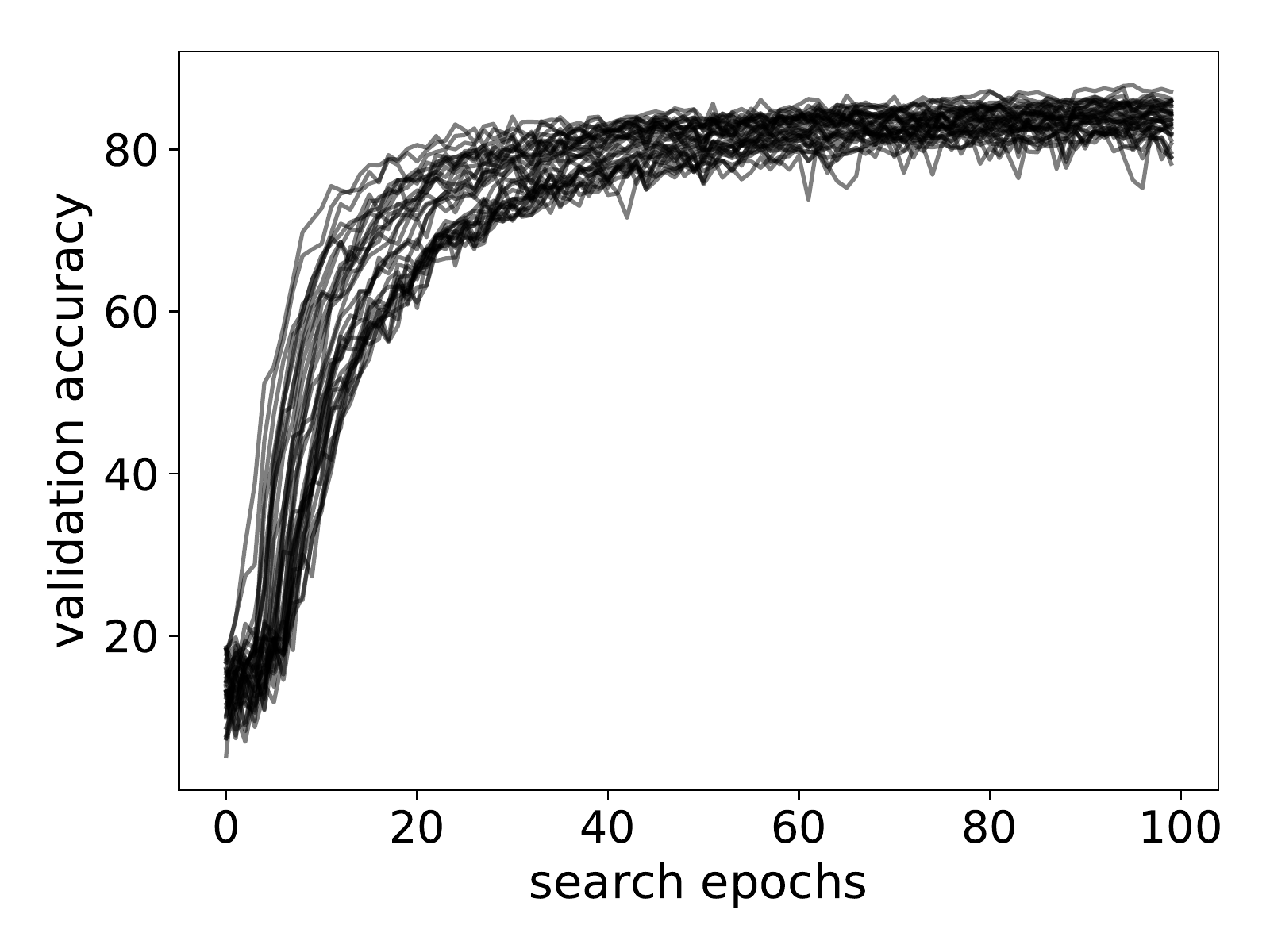}}\quad
  \subfigure[Entropy of the architecture distribution vs search epochs]{\includegraphics[width=.43\linewidth]{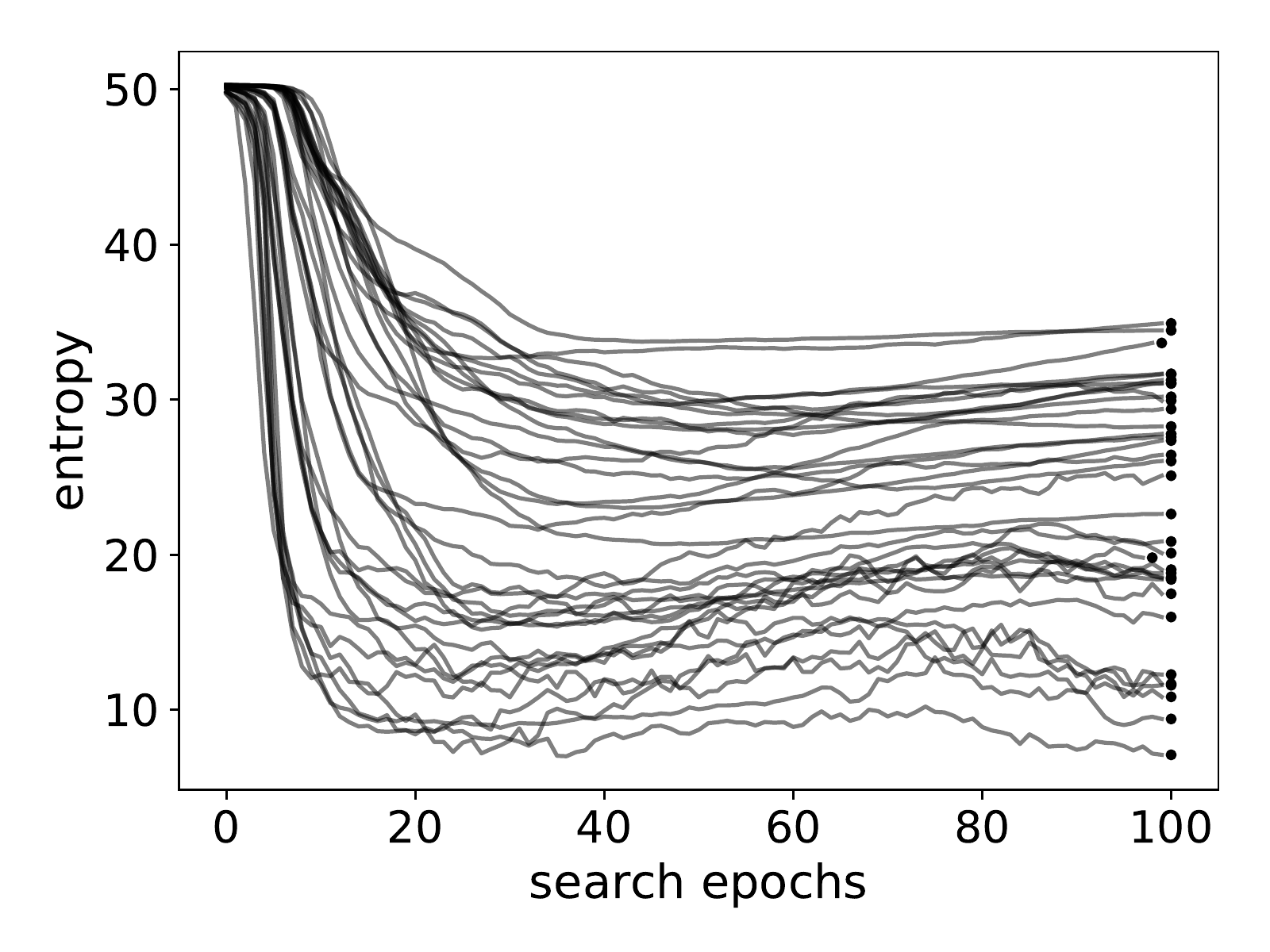}}
\caption{Validation accuracy of the mode of the architecture distribution and its entropy during independent architecture searches on CIFAR-10.}
\label{fig:search}
\end{figure*}

Figure \ref{fig:search}a shows the validation accuracy of the mode of the architecture distributions during search as a function of the number of epochs across multiple random seeds, learning rates and numbers of Monte Carlo samples $K$ (we considered $K=8, 16$, see also Section \ref{sec:iw-mceb}).
Figure\ref{fig:search}b shows the entropy of $p(\B{\alpha}|\B{\pi})$ as a function of the number of epochs in the same settings.
Overall, these figures show that the entropy of the architecture probability distribution steadily decreases over time (Figure \ref{fig:search}b) and that the mode of the architecture distributions (\ie the network we ultimately select) improves during search.

As reported in \citep{liu2018darts}, we also noticed variance in search results arising from subtle differences in initialization. For this reason, we run searches with 2 random seeds and select the architecture that reached highest validation accuracy. Search takes approximately 15 hours on a single V100 GPU.

To evaluate the accuracy of the resulting network, we train a large network of 20 cells. All the hyperparameters follow \citet{liu2018darts}. To summarize briefly: we train a network of 20 cells for 600 epochs and consider a batch size of 96. We use cutout in the data augmentation, consider a path dropout probability of 0.4 and auxiliary towers with weight 0.4. Since networks evaluated on CIFAR-10 exhibit high variance across random seeds \citep{liu2017hierarchical}, we report the mean and standard deviation across 5 random weight initializations.

\begin{table*}[!h]
  \centering
\resizebox{\textwidth}{!}{
  \begin{tabular}{@{}lccccc@{}}
  \toprule
  \multirow{2}{*}{\textbf{Architecture}} & \textbf{Test Error}      & \textbf{Params} & \textbf{Search Cost} & \textbf{Search} \\
  & \textbf{(\%)}      & \textbf{(M)} & \textbf{(GPU days)}  & \textbf{Method} \\ \midrule
DenseNet-BC \citep{huang2017densely}     & 3.46    & 25.6 & - & manual \\ \midrule
NASNet-A + cutout \citep{zoph2017learning}  & 2.65    & 3.3 & 1800 & RL \\
NASNet-A + cutout \citep{zoph2017learning}$^\dagger$ & 2.83    & 3.1 & 3150 & RL \\ 
AmoebaNet-A + cutout \citep{real2018regularized}  & 3.34 $\pm$ 0.06    & 3.2 & 3150 & evolution \\ 
AmoebaNet-A + cutout \citep{real2018regularized}$^\dagger$ & 3.12    & 3.1 & 3150 & evolution \\
AmoebaNet-B + cutout \citep{real2018regularized} & 2.55 $\pm$ 0.05 & 2.8 & 3150 & evolution \\
Hierarchical Evo \citep{liu2017hierarchical} & 3.75 $\pm$ 0.12 & 15.7 & 300 & evolution \\
PNAS \citep{liu2017progressive} & 3.41 $\pm$ 0.09 & 3.2 & 225 & SMBO \\
ENAS + cutout \citep{pham2018efficient} & 2.89    & 4.6 & 0.5 & RL \\ 
DARTS (first order)  + cutout \citep{liu2018darts} & 2.94 & 2.9 & 1.5 & gradient-based \\
DARTS (second order) + cutout \citep{liu2018darts} & 2.83 $\pm$ 0.06 &  3.4  & 4 & gradient-based \\ 
SNAS + mild constraint + cutout \citep{xie2018snas} & 2.98 &  2.9  & 1.5 & gradient-based \\ 
SNAS + moderate constraint + cutout \citep{xie2018snas} & 2.85 $\pm$ 0.02 &  2.8  & 1.5 & gradient-based \\ 
SNAS + aggressive constraint + cutout \citep{xie2018snas} & 3.10 $\pm$ 0.04 &  2.3  & 1.5 & gradient-based \\ 
\midrule
Random + cutout & 3.49 & 3.1      &    -     & -       \\ 
\textbf{\name (search on small network) + cutout} & 2.86 $\pm$ 0.06 & 3.6 & 0.6 & gradient-based \\ 
\textbf{\name (fine-tuning on large network) + cutout} & 2.81 $\pm$ 0.03 & 3.7 & 1 & gradient-based \\\bottomrule
  \end{tabular}}
  \caption{Comparison of different search methods and state-of-the-art architectures on CIFAR-10}
  \label{tab:cifar}
\end{table*}

As shown in table \ref{tab:cifar}, \name achieves results that are comparable to the state-of-the-art \citep{liu2018darts, xie2018snas} while requiring half of the computational cost.

\begin{figure*}
	\begin{center}
		\includegraphics[width=0.70\textwidth]{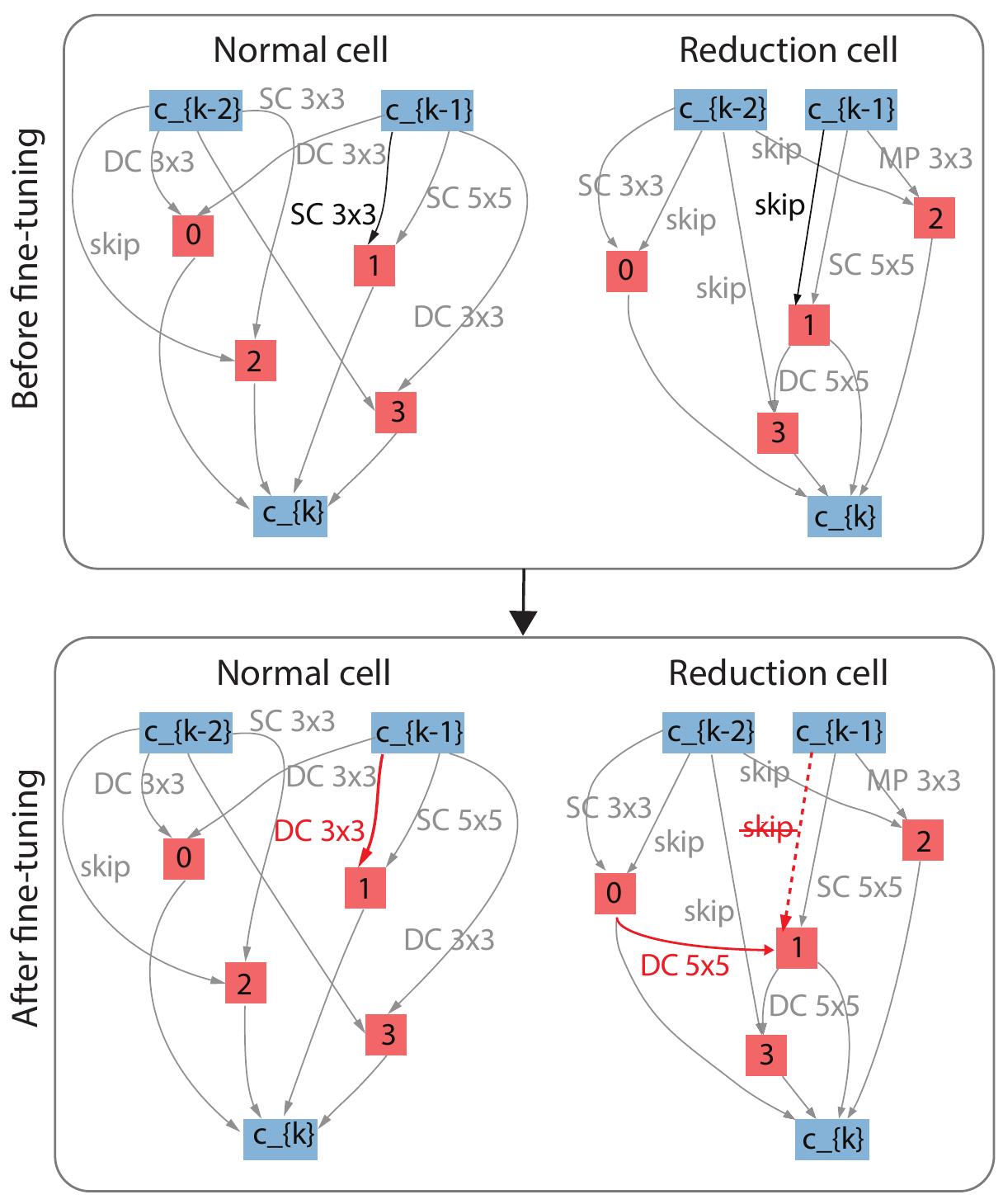}
	\end{center}
\caption{
Changes in the architecture in the fine-tuning experiment in CIFAR-10.
(top) Best architecture found on CIFAR-10 with the small network search.
(bottom) Best architecture after fine-tuning using the fully-sized network on CIFAR-10. Dashed lines represent connections that were removed.}
\label{fig:finetune}
\end{figure*}

\subsection{Architecture fine-tuning using a fully-sized network}
\label{sec:results_cifar_big}
Next, we transfer the distribution over architectures identified during the search on small networks and we tune it directly on fully-sized networks, removing the need for architecture surrogates entirely. This experiments showcases (i) the memory efficiency of our method, (ii) its computational efficiency and (iii) its ability to transfer knowledge across searches on different architectures.
For this fine-tuning task, we use an Adam learning rate of 0.01 and train for 10 epochs. The distribution $p(\mathbf{\alpha}|\mathbf{\pi})$ is initialized using the distribution tuned during the search described in the previous section. Importantly, we fine-tune on the exact network configuration that will be used to train the final network. This includes the use of auxiliary towers, cutout and larger batch sizes (96 samples instead of 64), which are all settings not considered during search on small networks.

This additional tuning step takes 8 hours on a single V100 GPU, bringing the total time taken by both search and fine-tuning to 23 hours.
Once more, we evaluate the accuracy of the resulting network by
training from scratch, as done in \citet{liu2018darts}.
As shown in table \ref{tab:cifar}, in exchange for a modest increase in search cost, fine-tuning further improves the accuracy of the networks identified during first search. While the results have relatively high variance, it's worth noting that on average, \name is more accurate than every method with comparable search cost, and it is only outperformed by methods \citep{zoph2017learning, real2018regularized} that have search budgets of thousands of GPU-days.

\subsection{Architecture evaluation on ImageNet}
\label{sec:results_imagenet}
\begin{table*}
	\centering
 \resizebox{\textwidth}{!}{
	\begin{tabular}{@{}lccccc@{}}
		\toprule
		\multirow{2}{*}{\textbf{Architecture}} & \multicolumn{2}{c}{\textbf{Test Error (\%)}}      & \textbf{Params} & \textbf{Search} & \textbf{Search} \\ \cline{2-3}
		& top-1 & top-5      & \textbf{(M)} &  \textbf{Cost} & \textbf{Method} \\ \midrule
		Inception-v1 \citep{szegedy2015going} & 30.2 & 10.1 & 6.6 & - & manual \\
		MobileNet \citep{howard2017mobilenets} & 29.4 & 10.5 & 4.2 & - & manual \\ 
		ShuffleNet 2$\times$ (v1) \citep{zhang2017shufflenet} & 29.1 & 10.2 & $\sim$5  & - & manual \\
		ShuffleNet 2$\times$ (v2) \citep{zhang2017shufflenet} & 26.3 & - & $\sim$5 & - & manual \\ \midrule
		NASNet-A \citep{zoph2017learning} & 26.0 & 8.4 & 5.3 & 1800 & RL \\
		NASNet-B \citep{zoph2017learning} & 27.2 & 8.7 & 5.3 & 1800 & RL \\
		NASNet-C \citep{zoph2017learning} & 27.5 & 9.0 & 4.9 & 1800 & RL \\
		AmoebaNet-A \citep{real2018regularized} & 25.5 & 8.0 & 5.1 & 3150 & evolution \\
		AmoebaNet-B \citep{real2018regularized} & 26.0 & 8.5 & 5.3 & 3150 & evolution \\
		AmoebaNet-C \citep{real2018regularized} & 24.3 & 7.6 & 6.4 & 3150 & evolution \\
		PNAS \citep{liu2017progressive} & 25.8 & 8.1 & 5.1 & ~225 & SMBO \\ 
		DARTS \citep{liu2018darts} &  26.9   &  9.0 & 4.9 & 4 & gradient-based \\ 
		SNAS + mild constraint \citep{xie2018snas} & 27.3 &  9.2  & 4.3 & 1.5 & gradient-based \\  \midrule
		\textbf{\name (search small network on CIFAR10)} &  26.3 &  8.4 & 5.5 & 0.6 & gradient-based \\
		\textbf{\name (fine-tuning large network on CIFAR10)} &  26.0 &  8.4 & 5.6 & 1 & gradient-based \\
		\bottomrule
	\end{tabular}
}
	\caption{Comparison with state-of-the-art image classifiers on ImageNet.}
	\label{tab:imagenet-results}
\end{table*}

Next, we transfer the architectures searched and fine-tuned on CIFAR-10 to ImageNet and train a large network to evaluate their performance. Once more, we follow the same hyperparameters used in \citet{liu2018darts} and train networks of 14 cells for 600 epochs.

As shown in Table \ref{tab:imagenet-results}, both architectures (searched on small network and fine-tuned on large network) identified by \name outperform those identified by competing methods with $1.5\times$ to $4\times$ larger computational costs. In particular the fine-tuned architecture also approaches the performance of the architecture identified by PNAS \citep{liu2017progressive}, which has a $225\times$ larger computational cost and matches the performance of NASNet \citep{zoph2017learning}, which has a $1800\times$ larger cost.

As observed in CIFAR-10, the fine-tuned architecture outperforms the architecture identified by searching over small architecture surrogates.

\section{Conclusion}
\label{sec:conclusions}

In this work, we have presented \name, a probabilistic approach neural architecture search method with low memory requirements that learns a probability distribution over high-performing neural network architectures.
Because of its efficient sampling-based optimization method, \name only requires as much memory as is needed to train a single architecture from its search space.
In our framework,
the architecture distributions learnt for a specific network and dataset can be transferred and fine-tuned in larger problems.
Through experiments on CIFAR-10 and ImageNet, we have shown that when searching over small network surrogates, our method matches or exceeds the performance of methods with same search space and comparable computational cost.
Importantly, we have shown that fine-tuning the architecture distribution on fully-sized network further improves performance on both CIFAR-10 and ImageNet. Our best network achieves a test error of 26\% on ImageNet, which is comparable to accuracies obtained with search algorithms that have computational costs that are up to three orders of magnitude larger.

\clearpage

\bibliography{references}
\bibliographystyle{abbrvnat}

\end{document}